\documentclass[runningheads]{llncs}
\usepackage{graphicx}
\usepackage{CJK}
\usepackage{nimbusmono}
\usepackage{booktabs}
\usepackage{multirow}
\usepackage[T1]{fontenc}
\begin{document}
\title{Improving the Generalization Ability in Essay Coherence Evaluation through Monotonic Constraints}
\titlerunning{Improving the Generalization in Essay Coherence Evaluation}
%
\author{Chen Zheng\inst{1,2} \and
Huan Zhang\inst{1} \and
Yan Zhao\inst{1} \and
Yuxuan Lai\inst{1,2}}

\authorrunning{C. Zheng et al.}

%
\institute{The Open University of China, Beijing, China \and
Engineering Research Center of Integration and Application of Digital Learning Technology, Ministry of Education, Beijing, China\\
\email{\{zhengchen,zhanghuan,zhaoyan,laiyx\}@ouchn.edu.cn}}
\maketitle              
\begin{abstract}
Coherence is a crucial aspect of evaluating text readability and can be assessed through two primary factors when evaluating an essay in a scoring scenario. The first factor is logical coherence, characterized by the appropriate use of discourse connectives and the establishment of logical relationships between sentences. The second factor is the appropriateness of punctuation, as inappropriate punctuation can lead to confused sentence structure. To address these concerns, we propose a coherence scoring model consisting of a regression model with two feature extractors: a local coherence discriminative model and a punctuation correction model. We employ gradient-boosting regression trees as the regression model and impose monotonicity constraints on the input features. The results show that our proposed model better generalizes unseen data. The model achieved third place in track 1 of NLPCC 2023 shared task 7. Additionally, we briefly introduce our solution for the remaining tracks, which achieves second place for track 2 and first place for both track 3 and track 4.

\keywords{Automated Essay Scoring \and Discourse Coherence \and Monotonic Constraints.}
\end{abstract}
\section{Introduction}
Discourse coherence refers to the degree to which the various components of a discourse are logically interconnected and contribute to a clear and meaningful message~\cite{ref_book1}. Analyzing coherence can greatly benefit numerous natural language processing tasks, such as text generation~\cite{ref_Huang},  summarization~\cite{ref_Christensen} and essay scoring~\cite{ref_Miltsakaki,ref_Burstein}.

In essay scoring tasks, there are many dimensions to measure the student's language proficiency, such as lexical sophistication, grammatical errors, content coverage and discourse coherence~\cite{ref_Cahill}. Since coherence is a key property of a well-written essay, coherence assessment plays an essential role in the task. 

In this work, we argue that two key aspects should be considered when evaluating the coherence of an essay. The first aspect is the logical coherence between sentences. The content of the essay should demonstrate a clear progression of ideas, with sentences and paragraphs closely connected and unfolding in logical order. Factors that may negatively impact the logical coherence between sentences include the improper use of discourse connectives and a lack of logical relationships between contexts. The second aspect is the appropriateness of punctuation. Proper punctuation is essential for clarifying the structure and organization of the essay. It can help establish logical connections between sentences, making the text easier to understand. Inappropriate punctuation can lead to confusion and disrupt the smooth flow of the text. 

In this work, we propose a feature-based coherence-scoring model framework. We employ two feature extractors to tackle the two essential aspects of coherence. Specifically, the first feature extractor is a local discriminative model~\cite{ref_Xu}, while the second is a punctuation correction model~\cite{ref_Zhang}. The local discriminative model takes two or three consecutive sentences as input and generates a probability estimate of the local coherence of the sequence. We separated the essay into successive sentences, taking each one as input for the model. Following the inference, we obtained the ratio of coherent sequences to the total number of sequences. The punctuation correction model examines the essay's punctuation usage and explicitly focuses on identifying redundant, missing, and misused commas and periods.

Following feature extractors, we propose employing a regression model to map features onto a final global coherence score. A simple yet transparent model for combining features is linear regression. However, when the patterns in the data exhibit non-linear relationships, alternative models such as random forest regression, gradient-boosted regression trees (GBRT), and neural networks offer superior performance compared to linear regression. A non-linear model may be prone to overfitting the data and negatively impacting the validity of automated scores. To address this issue, we enforce regulations on the input features to maintain linguistically-informed monotonicity, thereby enhancing scoring transparency and improving the model's generalization ability. 

Consequently, we present a scoring model that utilizes GBRT and incorporates monotonic constraints on the input features. We assume that the input feature, the ratio of locally coherent sequences to the total sequence of the essay, demonstrates a positive correlation with global coherence. Thus, we apply an increasing constraint to this feature. Furthermore, we assume that the feature of the number of redundant, missing, and misused commas and periods negatively correlates with global coherence. Hence, we impose a decreasing constraint on these features.

In summary, our contributions are as follows:
\begin{itemize}
\item We proposed a novel coherence scoring model consisting of a scorer with two feature extractors, i.e. a local discriminative model and a punctuation correction model. We showed that a local discriminative model with a more extended contextual input performs better than just consecutive pairs of sentences on the subsequent scoring tasks.    
\item We implement linguistically-informed monotonicity constraints on the input features to enhance the generalization ability in scoring essay coherence.
\item Experiments on the LEssay dataset demonstrate the effectiveness of our proposed methods, and we achieved third place on track 1 from NLPCC2023 shared task 7.
\end{itemize}
In the last of this paper, we will briefly overview our solution for the remaining tracks from NLPCC 2023 shared task 7. The code is available at\\
\url{https://github.com/chernzheng/nlpcc2023\_shared\_task7\_ouchnai\_solutions}.

\section{Related Works}
\subsubsection{Coherence Modeling} The early development of models for coherence analysis was influenced by lexical cohesion~\cite{ref_Halliday}, which refers to sharing identical or semantically related words in nearby sentences. Ref.~\cite{ref_Morris} introduced the concept of lexical chains and demonstrated that the number and density of lexical chains correlated with the topic structure. Ref.~\cite{ref_Hearst} introduced the TextTiling algorithm revealing that sentences or paragraphs within a subtopic exhibit higher cosine values than those in neighbouring subtopics. Ref.~\cite{ref_Foltz}'s LSA Coherence method pioneered the use of embeddings in studying coherence between sentences. 

Modern neural representation-learning coherence models~\cite{ref_Xu,ref_Li,ref_Mesgar} incorporate insights from early unsupervised coherence models for learning sentence representations and assessing their transformations between adjacent sentences. These models are designed to differentiate between natural and unnatural discourses based on deep neural networks. 

\subsubsection{Automated Chinese Essay Scoring} Ref.~\cite{ref_Zhang} implemented LDA to score Chinese essays. Ref.~\cite{ref_Fu} enhanced the accuracy of Chinese AES by recognizing beautiful sentences and incorporating them as literary features. Ref.~\cite{ref_Song1} assessed the organizational score of high school argumentative essays. Ref.~\cite{ref_Song2} investigated cross-prompt holistic scoring on four distinct essay sets, with articles in each dataset responding to a distinct prompt. Ref.~\cite{ref_He} proposed a multi-task learning framework for the Chinese AES and an inter-sequence attention mechanism to enhance information interaction between the different trait tasks.

\section{Method}

\begin{figure}[t]
\includegraphics[width=\textwidth]{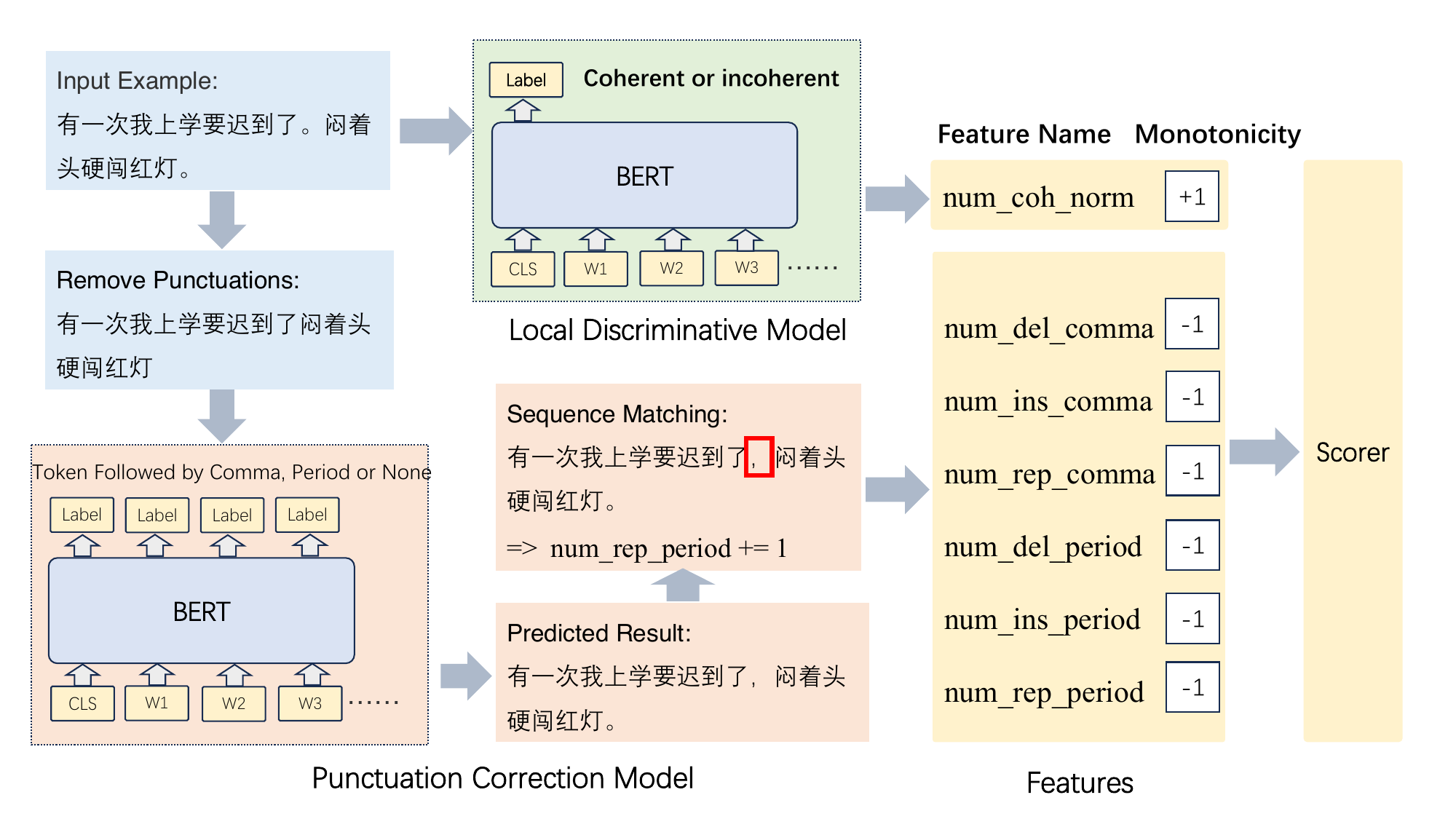}
\caption{The figure shows the architecture of our coherence scoring model. The punctuation correction model outputs six features: \texttt{num\_del\_comma}, \texttt{num\_ins\_comma}, \texttt{num\_rep\_comma}, \texttt{num\_del\_period}, \texttt{num\_ins\_period}, and \texttt{num\_rep\_period}, which enforced decreasing constraints on the subsequent scoring process. The local discriminative model output one feature: \texttt{num\_coh\_norm}, which enforces an increasing constraint.} \label{fig1}
\end{figure}

The architecture of our coherence scoring model is presented in Figure ~\ref{fig1}. The model consists of three components: a local discriminative model, a punctuation correction model, and a scorer. The local discriminative model is employed to evaluate the local coherence of consecutive sentences of the essay. The punctuation correction model is utilized to identify the inappropriateness of punctuation usage. The scorer maps the features extracted from the above two models into a final coherence score of the essay.

\subsection{Local Discriminative Model}

Our local discriminative model is similar to that of Ref.~\cite{ref_Xu}, but we employ BERT as an encoder and treat the problem as a text classification task. Ref.~\cite{ref_Xu} proposed a scoring model to differentiate between consecutive sentence pairs  in the training corpus, which are assumed to be coherent, and manually constructed incoherent ones. We extend the input sequence to three consecutive sentences rather than just two sentences and compare the different context lengths on the performance of subsequent scoring tasks.

For the case of sentence pairs, the input sequence is represented as \texttt{[CLS]} + Sentence \texttt{A} + \texttt{[SEP]} + Sentence \texttt{B}, where segment embeddings distinguish between the two sentences. For an essay with $n$ sentences, $s_{i}$ is the $i$-th sentence. We construct negative training samples by replacing one of the sentences, $s_{i}$ or $s_{i+1}$, with another sentence, $s_{j}$ ($j \neq i, i+1$), from the same essay. The trained model denoted as \textbf{LD-Bisent}). 

For the case of three sentences, the input sequence is set as \texttt{[CLS]} + Sentence \texttt{A} + Sentence \texttt{B} + Sentence \texttt{C} without using a special token \texttt{[SEP]} to separate them. We randomly substitute one sentence, $s_{i}$, $s_{i+1}$ or $s_{i+2}$, by $s_{j}$ ($j \neq i, i+1, i+2$) from the same essay as the negative training sample. The trained model denoted as \textbf{LD-Trisent}).

The model use the final hidden vector $C \in R^{H}$ (in our case, Chinese-RoBERTa-wwm-ext-large~\cite{ref_Cui}, H=1024) corresponding to the first input token \texttt{[CLS]} as the aggregate representation. The classification layer weights $W \in R^{K \times H}$, where K is the number of labels. In our case, $K=2$ for coherent or incoherent sequence. We compute a standard classification loss as $\log($\texttt{softmax}$(CW^{T}))$. 

\subsection{Punctuation Correction Model}

Our punctuation correction model is composed of two components. The first component, called the punctuation restoration model, accepts punctuation-free input texts and predicts the label for each token, indicating the punctuation that should follow it. The possible labels include a comma, a period, or no punctuation following the token. The second component is a misused-case classifier, which compares the punctuation-restored text with its original counterpart and determines the type of error the author has made. For instance, consider the sentence written by the author:

\begin{CJK*}{UTF8}{gkai}有一次我上学要迟到了。闷着头硬闯红灯。\end{CJK*}

\textit{(I ran late for school one day and recklessly charged through the red light.)}

\noindent To begin with, we remove the punctuation, resulting in the sentence

\begin{CJK*}{UTF8}{gkai}有一次我上学要迟到了闷着头硬闯红灯\end{CJK*}

\noindent Next, we input this sentence into the punctuation restoration model. The model predicts that the token `\begin{CJK*}{UTF8}{gkai}了\end{CJK*}' should be followed by a comma, the token `\begin{CJK*}{UTF8}{gkai}灯\end{CJK*}' should be followed by a period, and no punctuation following the rest of the token. Consequently, the punctuation-restored sentence becomes

\begin{CJK*}{UTF8}{gkai}有一次我上学要迟到了，闷着头硬闯红灯。\end{CJK*}

\noindent Subsequently, the misused-case classifier aligns the punctuation-restored sentence with its original counterpart and identifies that a comma has been erroneously used after the token `\begin{CJK*}{UTF8}{gkai}了\end{CJK*}'.

The punctuation restoration model is built upon a token classification model. We remove all punctuation marks from the original text and then pass it through a BERT encoder to obtain the final hidden vector for each input token $T_i \in R^H$. The probability of the token $i$ belonging to one of the labels \{0, 1, 2\} is computed as \texttt{softmax} $(S \cdot T_i)$, where $S \in R^{K \times H}$ is the set of weights to be learned of the final layer. Here, label 0 signifies that the token is not followed by punctuation, label 1 indicates a comma follows it, label 2 indicates it is followed by a period, and $K = 3$ is the number of labels.

The misused-case classifier uses a sequence-matching algorithm to compare the punctuation-restored texts with their original counterparts. We then count the instances of redundant, missing, and misused punctuation in the essay. For the sake of simplicity, all colons within the dataset are transformed into commas. Semicolons, question marks, and exclamation marks are replaced with periods while disregarding other punctuations. 

\subsection{Scorer}

The scorer takes extracted features from the above two models as input. The one feature is the ratio of coherent sequences to the total number of sequences in the essay (\texttt{num\_coh\_norm}). Additional features are the number of redundant, missing, and misused commas (\texttt{num\_del\_comma}, \texttt{num\_ins\_comma}, and \texttt{num\_rep\_comma}) and the period counterparts (\texttt{num\_del\_period}, \\
\texttt{num\_ins\_period}, and \texttt{num\_rep\_period}). 

We employ the abovementioned features as input and utilize a GBRT scorer with monotonic constraints to map these features into a final global coherence score. We impose a decreasing constraint for all features extracted from the punctuation correction model because these features characterize the inappropriateness of punctuation. For feature $x_i \in \{$\texttt{num\_del\_comma}, \\
\texttt{num\_ins\_comma},~~\texttt{num\_rep\_comma}, ~~\texttt{num\_del\_period}, \\
\texttt{num\_ins\_period},~~\texttt{num\_rep\_period}\}, the model satisfies
\begin{equation}
\mbox{GBRT}(x_1, \ldots, x_i, \ldots, x_n) \geq \mbox{GBRT}(x_1, \ldots, x'_i, \ldots, x_n)
\end{equation}
whenever $x_i \leq x'_i$. We impose an increasing constraint for feature $x_j =$\\
\texttt{num\_coh\_norm} because the feature captures the local coherence between adjacent sentences. It satisfies
\begin{equation}
\mbox{GBRT}(x_1, \ldots, x_j, \ldots, x_n) \leq \mbox{GBRT}(x_1, \ldots, x'_j, \ldots, x_n)
\end{equation}
whenever $x_j \leq x'_j$. 

We compare our proposed scoring model against two regression models: a linear model and a random forest model. We also compare the performance of our model with different configurations, i.e. the scorer with or without monotonic constraints and the local discriminative model with different context lengths.

\section{Experiments}

\subsection{Datasets}

\subsubsection{LEssay dataset} The LEssay dataset consists of four sub-datasets corresponding to four tasks. All tasks are related to the coherence evaluation of Chinese student essays. The first sub-dataset is dedicated to the task of global coherence evaluation. It includes a training set of 50 essays, a verification set of 10 essays, and a test set of 5,000 essays. All of these essays are written in Chinese by middle school students and assessed for their coherence on three levels: excellent, moderate, and poor. The remaining three sub-datasets are allocated to the topic sentence extraction, paragraph and sentence logical relation recognition tasks, respectively. 

These four tasks are interconnected, and a model trained on one sub-dataset can potentially contribute to another task. However, in this study, a global coherence scoring model will be trained only by the first sub-dataset and two external datasets. These external datasets, including the Chinese essay dataset for pre-training~\cite{ref_Song2} and the IWSLT 2012-zh dataset for punctuation restoration~\cite{ref_Federico}, will be utilized to train the feature extractors for the scoring model. The global coherence scores of the first sub-dataset will be used to train the scorer.

\subsubsection{Chinese essay dataset for pre-training} The dataset comprises 93,002 essays authored by Chinese students in grades 7 to 12, covering various topics and genres, such as narrative, argumentative, and expository essays.

We utilized the dataset for training the local discriminative model. In practice, we excluded essays with the lowest rating (assigned rating 1) due to poor writing quality. For the remaining essays, we divided each into consecutive sentence pairs or triple sentences, assuming their coherence. And we manually created incoherent sentences, as described in section 3.1. We generated 4.3 million positive and equal negative training samples for the LD-Bisent. We also prepared 3.1 million positive and equal negative training samples for the LD-Trisent.

\subsubsection{IWSLT2012-zh dataset} The dataset consists of 150k lines of sentences in Chinese from TED talk transcripts. We only predict commas and periods. The question marks are converted to periods for simplicity.

\subsection{Experimental Settings}

We use the pre-trained Chinese-RoBERTa-wwm-ext-large model to fine-tune the local discriminative and punctuation correction models. For the random forest scorer, we set the number of trees in the forest to 30 and maintained the other parameters at their default values. For the GBRT scorers, we configure the number of boosted gradients to 30, with a maximum tree depth for base learners of 4. The learning rate is set to 1, and all other parameters are left at their default values. 

We use precision, recall, and macro F1-score to evaluate the effectiveness of coherence identification. The precision is calculated by dividing the number of correctly identified coherence types (excellent, moderate, and poor) by the total number of identified coherence types. The recall is determined by dividing the number of correctly identified coherence types by the total number of coherence types as labelled.

\begin{table}
\centering
\caption{Comparison of regression models}\label{tab1}
\begin{tabular}{lccc}
\toprule
\textbf{ Model } & \textbf{ Precision } & \textbf{ Recall } & \textbf{ Macro F1 }\\
\midrule
 Linear Regression & 35.55 & 48.44 & 25.57\\
 Random Forest Regression & 38.86 & 23.44 & 28.74\\
 \midrule
 GBRT (Bi-sent) & 33.41 & 34.10 & 31.82\\
 GBRT w/ MC (Bi-sent) & 36.98 & 23.02 & 26.67\\
 \midrule
 GBRT (Tri-sent) & 35.77 & 36.26 & 34.52\\
 GBRT w/ MC (Tri-sent) & 37.28 & 39.90 & 33.02\\
\bottomrule
\end{tabular}
\end{table}

\subsection{Results}
Table~\ref{tab1} presents the results of each regression model. In the experiment, we used the LD-Trisent feature extractor in linear and random forest regressions.

Our findings suggest that the GBRT model with monotonic constraint using LD-Trisent (GBRT w/ MC (Tri-sent)) performs better in terms of precision and recall compared to the same model without enforcing monotonic constraint (GBRT (Tri-sent)). Furthermore, this model demonstrates improvements in precision, recall, and macro F1 score compared to the same model using LD-Bisent (GBRT w/ MC (Bi-sent)) and LD-Bisent without enforcing monotonic constraint (GBRT (Bi-sent)). Additionally, this model exhibits superior performance in macro F1 score compared to both linear and random forest regressions.

Our results show that training local coherence models to predict longer contexts than just consecutive pairs of sentences can result in better performance on subsequent scoring tasks, which agrees with the previous study on discourse representation~\cite{ref_Iter}.

\section{Our Solution to the Remaining Tracks from NLPCC2023 shared task7}
\subsection{Text Topic Extraction (Track 2)}

This task aims to identify the topic sentence for each paragraph and one overall topic sentence for a given middle school student essay. 

In our approach, we employ two token classification models to identify both paragraph-level and overall topic sentences.
The first model accepts the essay title connected to a paragraph as input. For each token, it outputs a label indicating whether the token belongs to the topic sentences of the paragraph (designated as a key token). The topic sentences of each paragraph are determined by the ratio of key tokens to the total number of tokens within the sentence. We select the sentence with the highest ratio as the topic sentence for that paragraph. The model is fine-tuned on Chinese-RoBERTa-wwm-ext-large.

The second model is similar to the first, but the input is a sequence that sequentially connects the essay title to all paragraph's topic sentences. We assume that the overall topic sentence is one of the paragraph topic sentences and determine it by calculating the ratio of key tokens to the total number of tokens within each paragraph topic sentence. We select the sentence with the highest ratio as the overall topic sentence. The second model is fine-tuned on the first model.

The evaluation results are shown in Table~\ref{tab2}. Our approach achieved second place in Track 2.
\begin{table}
\centering
\caption{The result of text topic extraction.}\label{tab2}
\begin{tabular}{lccccc}
\toprule
\textbf{Team} & \textbf{ Para. Acc. } & \textbf{ Full Acc. } & \textbf{ Final Acc. } & \textbf{ Para. Simi. } & \textbf{ Full Simi. }\\
\midrule
wuwuwu & 61.27 & 34.92 & 42.82 & 87.34 & 80.37\\
\textbf{Ours} & 62.61 & 33.33 & 42.12 & 85.20 & 79.16\\
\bottomrule
\end{tabular}
\end{table}

\subsection{Paragraph Logical Relation Recognition (Track 3)}
The task aims to determine the logical relationship between the two consecutive paragraphs of an essay. The logical relationship includes co-occurrence, inversion, explanatory and superior-subordinate relationships.

Our approach regards the paragraph-level logical relation recognition task as a sequence classification problem. Specifically, we process a pair of paragraphs as input, and the model determines the logical relationship between these paragraphs. Considering the similarity between this task and sentence-level logical relation recognition, we chose to fine-tune the model trained for track 4.

The evaluation results for track 3 are shown in Table~\ref{tab3}. Our approach achieved first place in the track. 

\begin{table}
\centering
\caption{The results of paragraph-level logical relation recognition.}\label{tab3}
\begin{tabular}{lccc}
\toprule
\textbf{ Team } & \textbf{ Precision } & \textbf{ Recall } & \textbf{ Macro F1 }\\
\midrule
 \textbf{Ours} & 54.66 & 52.45 & 52.16 \\
 wuwuwu & 29.26 & 28.98 & 28.77 \\
 Lrt123 & 28.19 & 30.26 & 27.54\\
 BLCU\_teamworkers & 27.17 & 27.65 & 25.95\\
\bottomrule
\end{tabular}
\end{table}
\subsection{Sentence Logical Relation Recognition (Track 4)}
The task is comparable to the previous task. Nonetheless, the logical relationships are sentence-based and include 12 different relationships.

We employ a two-stage training approach for our classification model. In the first stage, we utilize an external dataset, TED-CDB~\cite{ref_Long}, to pre-train the model based on Chinese-RoBERTa-wwm-ext-large. In the subsequent stage, we fine-tune the pre-trained model on the current dataset to enhance its performance for the given task.

The evaluation results for track 4 are shown in Table~\ref{tab4}. Our approach achieved first place in the track. 
\begin{table}
\centering
\caption{The results of sentence-level logical relation recognition.}\label{tab4}
\begin{tabular}{lccc}
\toprule
\textbf{ Team } & \textbf{ Precision } & \textbf{ Recall } & \textbf{ Macro F1 }\\
\midrule
 \textbf{Ours} & 36.63 & 36.36 & 34.38 \\
 wuwuwu & 23.49 & 25.37 & 23.67 \\
 BLCU\_teamworkers & 7.55 & 6.30 & 6.32 \\
\bottomrule
\end{tabular}
\end{table}

\section{Conclusion and Future Work}
In this study, we present a scoring model to assess the global coherence of Chinese student essays. This scoring model incorporates two feature extractors: a local coherence discriminative model and a punctuation correction model. Furthermore, we employed a GBRT model with linguistically-informed monotonicity constraints to convert features into a final global coherence score.

Our findings suggest that the enforced regulations on the features improved the model's generalization capability, and a local discriminative model with a context extending beyond consecutive sentence pairs can achieve better performance in scoring tasks.

For future research, we will incorporate the features of paragraph-level coherence into the scoring model. The current model considers sentence-level coherence by introducing a local discriminative model. But the global coherence characterized by logical relationships between paragraphs is equally important for coherence evaluation. By incorporating paragraph-level coherence features, we can further enhance the performance of the scoring model and provide a more accurate assessment.

\section{Acknowledgements}
This work is supported by NSFC (62206070), the Innovation Fund Project of the Engineering Research Center of Integration and Application of Digital Learning Technology, Ministry of Education (1221014, 1221052), and National Key R\&D Program of China (2021YFF0901005).

%
%
%
%

\end{document}